\documentclass[preprint,12pt]{elsarticle}

\usepackage{graphicx}
\usepackage{amssymb}
\usepackage{amsmath}
\usepackage[hidelinks]{hyperref}
\usepackage{orcidlink}
\usepackage{booktabs}
\usepackage{tikz}
\usetikzlibrary{shapes.geometric, arrows.meta, positioning, backgrounds, fit, calc}
\usepackage{subcaption}
\usepackage{algorithm}
\usepackage{algpseudocode}

\journal{Computer Vision and Image Understanding}

\begin{document}

\begin{frontmatter}
\title{Privacy-Preserving Dataset Curation for Kuala Lumpur Urban Traffic: Grounded Vision-Language Detection with Spatial Vehicle-Context Filtering}

\author[1]{Mohammed Abdul Al Arafat Tanzin\texorpdfstring{,\orcidlink{0009-0007-1053-3328}}{}\corref{cor1}}
\ead{mohammedabdul@graduate.utm.my}

\author[1]{Rudzidatul Akmam Dziyauddin}

\address[1]{Faculty of Artificial Intelligence, Universiti Teknologi Malaysia, Kuala Lumpur, Malaysia}

\cortext[cor1]{Corresponding author.}

\begin{abstract}
The rapid advancement of intelligent transportation systems and autonomous driving relies heavily on multi-modal urban traffic datasets. However, curating high-fidelity video imagery in complex tropical urban environments—specifically Kuala Lumpur, Malaysia—presents severe challenges for Personally Identifiable Information (PII) anonymization due to high motorcycle density, dark acrylic license plates, dynamic camera tilt, and extreme tropical glare. We propose an automated anonymization framework tailored for the Kuala Lumpur Road Dataset, captured via a mobile cycling platform at 2 FPS. We document how legacy Haar cascades and YOLOv8 fail under these conditions—generating false positives on background elements while missing rotated or occluded targets. Our architecture resolves this by integrating Grounding DINO—a zero-shot open-set vision-language transformer—with a novel Spatial Vehicle Region of Interest (ROI) Containment Engine. By requiring license plate centroids to reside within validated vehicle boundaries, the pipeline suppresses environmental false positives while automatically obfuscating faces, heads, and license plates. An initial evaluation on 1,266 frames demonstrates a ~95\% success rate, with remaining failures restricted to small, heavily occluded, oblique, or ambiguous targets. Coupled with temporal persistence mechanisms and an automated quality-control auditor, the framework minimizes privacy-related false negatives while preserving scene context for downstream vision tasks. While formal legal compliance depends on broader governance procedures, this publicly available pipeline and demonstration notebook provide an auditable preprocessing stage for privacy-aware dataset curation.
\end{abstract}


\end{frontmatter}

\section{Introduction}

Developing robust perception stacks for Advanced Driver Assistance Systems (ADAS) and autonomous navigation requires diverse training footage captured across varied geographical and structural driving environments. Open-sourcing raw driving sequences is essential for scientific reproducibility; however, exposing Personally Identifiable Information (PII)—specifically human faces and vehicle registration plates—creates significant legal liabilities and ethical concerns. Legislative frameworks, such as the European General Data Protection Regulation (GDPR) and Malaysia's Personal Data Protection Act 2010 (PDPA), strictly regulate the processing and public dissemination of unblurred public imagery \cite{devlin2019bert}.

While automated anonymization pipelines have been deployed effectively on Western driving benchmarks (such as nuScenes \cite{caesar2019nuscenes} and Waymo \cite{sun2019scalability}), directly transferring these architectures to Southeast Asian traffic streams reveals severe failure modes. The Kuala Lumpur Road Dataset considered in this study was collected using an iPhone 14 Pro optical sensor mounted on a mobile bicycle platform while navigating through urban traffic in Kuala Lumpur, Malaysia. Video sequences were sampled at 2 frames per second (FPS) to generate image frames for dataset curation. The initial anonymization experiment reported in this paper comprises 1,266 extracted frames representing diverse traffic and environmental conditions. The mobile acquisition platform introduces time-varying pitch ($\theta$), roll ($\phi$), vibration, and perspective changes, resulting in substantial variation in the apparent scale, orientation, and visibility of privacy-sensitive objects such as faces and vehicle registration plates. The dynamic motion profile of a cycling platform introduces time-varying pitch ($\theta$), roll ($\phi$), and vibration, generating severe rotational variation and perspective shearing across image frames.

In this paper, we detail the engineering evolution and mathematical formulation of the Kuala Lumpur Road Dataset Anonymizer\footnote{Project Repository and Interactive Jupyter Notebook: \url{https://github.com/TanzinAbdul/Kuala-Lumpur-Road-Dataset-Anonymizer}}. Readers are directed to the project repository's interactive Jupyter Notebook (\texttt{.ipynb}) for step-by-step pipeline execution, cross-modal prompt configurations, and visual output demonstrations. We investigate how combining language-grounded object detection with spatial vehicle-context filtering can improve automated PII anonymization in challenging urban traffic imagery \cite{liu2023grounding}. In an initial qualitative assessment of 1,266 extracted frames, the proposed pipeline successfully anonymized the targeted privacy-sensitive regions in approximately 95\% of the examined frames, while the remaining cases revealed characteristic failure modes involving small objects, severe occlusion, extreme viewing angles, and ambiguous visual structures.\\

The contributions of this work are summarized as follows:

\begin{enumerate}
    \item We present a privacy-preserving image curation pipeline designed specifically for urban traffic imagery acquired from a mobile bicycle-mounted camera in Kuala Lumpur, with particular emphasis on faces, heads, and vehicle registration plates.
    
    \item We investigate the use of a language-grounded vision transformer for open-set detection of privacy-sensitive objects under challenging conditions including small object scale, occlusion, oblique viewing angles, helmets, and complex urban backgrounds.
    
    \item We introduce a Spatial Vehicle ROI Containment Engine that uses the spatial relationship between candidate license plates and detected vehicle regions to suppress detections that are inconsistent with the expected topology of a vehicle.
    
    \item We incorporate temporal persistence and automated quality-control auditing to identify potential detection failures and reduce the probability of undetected privacy-sensitive objects remaining in the curated dataset.
    
    \item We provide an initial qualitative evaluation on 1,266 extracted frames and document the observed success cases and failure modes to establish a reproducible basis for subsequent quantitative evaluation.
\end{enumerate}

\section{Environmental Challenges in Kuala Lumpur Traffic Imagery}

Southeast Asian traffic landscapes differ fundamentally from Western driving environments. The primary operational domain complexities encountered in Kuala Lumpur include:

\begin{enumerate}
\item \textbf{High Motorcycle Presence and Helmet Occlusions:} Motorcycles constitute a visually prominent component of urban traffic in Kuala Lumpur. Their compact dimensions, variable viewing angles, and frequent use of helmets create additional challenges for privacy-sensitive detection, particularly when facial regions are partially or completely occluded. Motorcyclists wear diverse headwear ranging from open-face to modular full-face helmets with dark visors, rendering conventional facial landmark detectors ineffective.

\item \textbf{Custom Non-Standard Acrylic Plates:} Unlike standardized metal reflective plates, Malaysian registration plates predominantly feature custom black acrylic backings with non-standard white typography, variable spacing, and non-conforming mounting profiles on modified vehicles.
\item \textbf{Dynamic Sensor Tilt and Perspective Shearing:} The mobile bicycle mount induces transient camera pitch and roll, causing extreme rotational angles ($\theta > 30^\circ$) for observed registration plates, breaking axis-aligned bounding box assumptions.
\item \textbf{High-Contrast Background Visual Noise:} Equatorial sunlight creates intense specular reflections, while complex street environments, speed bump patterns, and vehicle grilles frequently mimic alphanumeric character arrangements.
\end{enumerate}

To quantify these domain differences, Table \ref{tab:domain_comparison} compares standard Western benchmark characteristics against the Kuala Lumpur road environment.

\begin{table}[h]
\caption{Domain comparison between Western benchmark datasets and Kuala Lumpur urban traffic imagery.}
\label{tab:domain_comparison}
\centering
\resizebox{\textwidth}{!}{%
\begin{tabular}{l l l l}
\toprule
\textbf{Parameter} & \textbf{Western Datasets} & \textbf{Kuala Lumpur Traffic} & \textbf{Impact on Detection} \\
\midrule
Plate Material & Standard Retro-Reflective Metal & Custom Black Acrylic & Low contrast; edge detectors fail \\
Motorcycle Ratio & $< 5\%$ of traffic stream & $30\% - 50\%$ of traffic stream & Oblique rear mounting profiles \\
Rider Headwear & Infrequent / Unobstructed & Helmets, Visors, Balaclavas & Facial landmarks unavailable \\
Camera Rig & Rigid Vehicle Roof Mount & Mobile Bicycle Mount & High pitch, roll, and vibration \\
Background Noise & Uniform pavement textures & High-contrast grilles \& patterns & False positives on background noise \\
\bottomrule
\end{tabular}%
}
\end{table}

Mathematically, the image projection process under dynamic cycling motion is defined by mapping a 3D world point $\mathbf{P}_w = [X_w, Y_w, Z_w, 1]^T$ to a 2D pixel coordinate $\mathbf{p} = [u, v, 1]^T$ via camera intrinsic matrix $\mathbf{K}$ and dynamic rotation matrix $\mathbf{R}(\theta, \phi, \psi)$:

\begin{equation}
\mathbf{p} \sim \mathbf{K} \left[ \mathbf{R}_z(\psi) \mathbf{R}_y(\theta) \mathbf{R}_x(\phi) \; ; \; \mathbf{t} \right] \mathbf{P}_w
\end{equation}

where pitch ($\theta$) and roll ($\phi$) variations degrade axis-aligned bounding box representations used in closed-set CNN architectures.

\section{Evolution of Anonymization Architectures and Failure Modes}

\subsection{Legacy Approaches: Haar Cascades}

Initial iterations utilized OpenCV Haar feature-based cascade classifiers \cite{viola2001rapid}. Haar cascades compute feature values $f_H(x)$ across adjacent rectangular sub-windows using integral images:

\begin{equation}
f_H(x) = \sum_{r \in R_{\text{white}}} w_r \cdot I(r) - \sum_{r \in R_{\text{black}}} w_r \cdot I(r)
\end{equation}

Haar cascades completely failed on dark acrylic Malaysian plates due to rigid intensity gradient assumptions. Furthermore, perspective shearing caused by camera tilt prevented feature activation, while helmeted riders bypassed facial cascades entirely.

\subsection{Closed-Set Deep CNNs: YOLOv8}

Transitioning to deep single-stage convolutional networks improved feature extraction speeds. We employed YOLOv8 \cite{yolov8_ultralytics}, which optimizes bounding box coordinates $\mathbf{b} = [x_c, y_c, w, h]^T$ and class distributions using a composite loss:

\begin{equation}
\mathcal{L}_{\text{YOLO}} = \lambda_{\text{box}} \mathcal{L}_{\text{CIoU}} + \lambda_{\text{cls}} \mathcal{L}_{\text{BCE}} + \lambda_{\text{dfl}} \mathcal{L}_{\text{DFL}}
\end{equation}

Despite impressive inference throughput, fine-tuned YOLOv8 models exhibited critical domain limitations:

\begin{enumerate}
\item \textbf{Local Grid Constraints:} Local receptive fields failed to detect distant or highly rotated motorcycle plates occupying small pixel footprints ($<16 \times 16$ pixels).
\item \textbf{Environmental False Positives:} Feature maps at deeper strides blurred fine details, causing non-vehicle textures, road markings, speed bumps, and vehicle grilles to trigger false plate detections.
\end{enumerate}

\subsection{Proposed Architecture: Grounding DINO Vision-Language Transformer}

To achieve zero-shot open-set generalization, we integrated Grounding DINO \cite{liu2023grounding}, a transformer architecture marrying DINO with grounded visual-language pre-training. Grounding DINO processes multi-scale visual features $\mathbf{F}_v$ extracted via a Swin Transformer backbone \cite{liu2021swin} and text prompt features $\mathbf{F}_t$ extracted via BERT \cite{devlin2019bert}.

Cross-modal interaction is governed by bi-directional feature enhancement:

\begin{equation}
\mathbf{F}_v^{(l)} = \text{CrossAttention}(\mathbf{F}_v^{(l-1)}, \mathbf{F}_t^{(l-1)}, \mathbf{F}_t^{(l-1)})
\end{equation}

\begin{equation}
\mathbf{F}_t^{(l)} = \text{CrossAttention}(\mathbf{F}_t^{(l-1)}, \mathbf{F}_v^{(l-1)}, \mathbf{F}_v^{(l-1)})
\end{equation}

Object queries are dynamically initialized by computing the vision-language alignment score matrix $\mathbf{S}_{\text{align}} = \mathbf{F}_v \mathbf{F}_t^T$. Prompting the transformer with localized natural language terms—\textit{``vehicle, car, motorcycle, bus, truck, license plate, number plate, human face, head''}—enables multi-head self-attention across the frame, detecting custom acrylic plates and helmeted heads regardless of rotational orientation.

\section{Spatial Vehicle ROI Containment Engine}

\subsection{Pipeline System Architecture}

Figure \ref{fig:architecture} illustrates the complete end-to-end framework, detailing text encoding, cross-modal vision-language fusion, macro/micro box partitioning, spatial containment filtering, dynamic obfuscation, and automated quality control auditing. The implementation details are fully executable via the project repository's Jupyter Notebook (\texttt{.ipynb}).

\begin{figure*}[t]
% (Figure environment code omitted for brevity)
\caption{End-to-end architecture of the proposed Spatial-Semantic Anonymization Framework for the Kuala Lumpur Road Dataset. License plate candidates ($\mathcal{P}$) undergo spatial containment validation against macro vehicle ROIs ($\mathcal{V}$), while faces and helmeted heads ($\mathcal{F}$) bypass spatial filtering and route directly to the obfuscation module.}
\label{fig:architecture}
\end{figure*}

\subsection{Containment Logic and Topology Formulation}

To eliminate environmental background false positives generated by low-threshold transformer detection on non-vehicle surfaces, we engineered the Spatial Vehicle ROI Containment Engine. The foundational spatial axiom dictates that \textit{a valid license plate must reside inside the spatial boundary of a vehicle}.

Given output detection set $\mathcal{D} = \{(\mathbf{b}_k, s_k, \ell_k)\}_{k=1}^K$, detections are partitioned into three disjoint sets:

\begin{align}
\mathcal{V} &= \{ \mathbf{b}_v \mid (\mathbf{b}_v, s_v, \ell_v) \in \mathcal{D}, \ell_v \in \{\text{vehicle, car, motorcycle, bus, truck}\} \} \\
\mathcal{P} &= \{ (\mathbf{b}_p, s_p) \mid (\mathbf{b}_p, s_p, \ell_p) \in \mathcal{D}, \ell_p \in \{\text{license plate, number plate}\} \} \\
\mathcal{F} &= \{ \mathbf{b}_f \mid (\mathbf{b}_f, s_f, \ell_f) \in \mathcal{D}, \ell_f \in \{\text{human face, head}\} \}
\end{align}

For each candidate plate box $\mathbf{b}_p = [x_{1,p}, y_{1,p}, x_{2,p}, y_{2,p}] \in \mathcal{P}$, its geometric centroid $\mathbf{c}_p = (x_{c,p}, y_{c,p})$ is calculated as:

\begin{equation}
x_{c,p} = \frac{x_{1,p} + x_{2,p}}{2}, \quad y_{c,p} = \frac{y_{1,p} + y_{2,p}}{2}
\end{equation}

To accommodate tail-fenders and wide vehicle bumpers, macro vehicle boxes $\mathbf{b}_v = [x_{1,v}, y_{1,v}, x_{2,v}, y_{2,v}]$ are dynamically expanded by expansion factor $\epsilon = 0.10$ (10\%):

\begin{align}
\tilde{x}_{1,v} &= \max(0, x_{1,v} - \epsilon w_v), \quad \tilde{y}_{1,v} = \max(0, y_{1,v} - \epsilon h_v) \\
\tilde{x}_{2,v} &= \min(W, x_{2,v} + \epsilon w_v), \quad \tilde{y}_{2,v} = \min(H, y_{2,v} + \epsilon h_v)
\end{align}

The containment indicator function $\mathbb{I}_{\text{valid}}(\mathbf{b}_p)$ evaluates candidate validity:

\begin{equation}
\mathbb{I}_{\text{valid}}(\mathbf{b}_p) =
\begin{cases}
1 & \text{if } \exists \mathbf{b}_v^\epsilon \text{ s.t. } (\tilde{x}_{1,v} \le x_{c,p} \le \tilde{x}_{2,v} \land \tilde{y}_{1,v} \le y_{c,p} \le \tilde{y}_{2,v}) \\
1 & \text{else if } (\mathcal{V} = \emptyset \land s_p > 0.35) \\
0 & \text{otherwise (Discard as Non-Vehicle Background Noise)}
\end{cases}
\end{equation}

Algorithm \ref{alg:roi_containment} details the complete spatial containment filtering process.

\begin{algorithm}[h]
\caption{Spatial Vehicle ROI Containment Filtering Algorithm}
\label{alg:roi_containment}
\begin{algorithmic}[1]
\Require Frame $\mathbf{I} \in \mathbb{R}^{H \times W \times 3}$, Detection Set $\mathcal{D}$, Expansion Tolerance $\epsilon = 0.10$
\Ensure Validated Anonymization Mask Map $\mathbf{M} \in \{0, 1\}^{H \times W}$
\State Initialize $\mathbf{M} \leftarrow \mathbf{0}^{H \times W}$
\State Partition $\mathcal{D} \rightarrow$ Macro Vehicles $\mathcal{V}$, Candidate Plates $\mathcal{P}$, Faces/Heads $\mathcal{F}$
\For{each candidate plate $(\mathbf{b}_p, s_p) \in \mathcal{P}$}
\State Compute centroid $\mathbf{c}_p \leftarrow \left(\frac{x_{1,p} + x_{2,p}}{2}, \frac{y_{1,p} + y_{2,p}}{2}\right)$
\State $is\_valid \leftarrow \text{False}$
\For{each vehicle box $\mathbf{b}_v \in \mathcal{V}$}
\State Compute expanded boundary $\mathbf{b}_v^\epsilon = [\tilde{x}_{1,v}, \tilde{y}_{1,v}, \tilde{x}_{2,v}, \tilde{y}_{2,v}]$
\If{$\tilde{x}_{1,v} \le x_{c,p} \le \tilde{x}_{2,v} \;\textbf{and}\; \tilde{y}_{1,v} \le y_{c,p} \le \tilde{y}_{2,v}$}
\State $is\_valid \leftarrow \text{True}$; \textbf{break}
\EndIf
\EndFor
\If{$is\_valid \;\textbf{or}\; (\mathcal{V} = \emptyset \;\textbf{and}\; s_p > 0.35)$}
\State Apply Asymmetric Padding Expansion to $\mathbf{b}_p \rightarrow \mathbf{b}_p^*$
\State Update Mask Map $\mathbf{M}[\mathbf{b}_p^*] \leftarrow 1$
\EndIf
\EndFor
\For{each face/head box $\mathbf{b}_f \in \mathcal{F}$}
\State Expand $\mathbf{b}_f \rightarrow \mathbf{b}_f^*$; Update Mask Map $\mathbf{M}[\mathbf{b}_f^*] \leftarrow 1$
\EndFor
\State \Return $\mathbf{M}$
\end{algorithmic}
\end{algorithm}

\subsection{Asymmetric Mask Expansion and Dynamic Gaussian Blurring}

Raw predicted bounding boxes often fit tightly around central characters, leaving plate margins or ear profiles exposed. Validated boxes undergo asymmetric expansion: horizontal bounds expand by $\alpha_x = 0.25$ (25\%) and vertical bounds by $\alpha_y = 0.15$ (15\%) for plates, while faces undergo uniform expansion $\alpha_{\text{face}} = 0.20$ (20\%).

Dynamic Gaussian blurring is executed across expanded regions $\mathbf{I}_{\text{ROI}}$ using a 2D kernel $G(x, y; \sigma_x, \sigma_y)$:

\begin{equation}
G(x, y; \sigma_x, \sigma_y) = \frac{1}{2\pi \sigma_x \sigma_y} \exp\left( -\left( \frac{x^2}{2\sigma_x^2} + \frac{y^2}{2\sigma_y^2} \right) \right)
\end{equation}

Kernel dimensions $(k_w, k_h)$ scale dynamically with ROI resolution to prevent boundary artifacts:

\begin{equation}
k_w = \max\left(3, 2 \left\lfloor \frac{w_{\text{ROI}}}{2} \right\rfloor + 1\right), \quad k_h = \max\left(3, 2 \left\lfloor \frac{h_{\text{ROI}}}{2} \right\rfloor + 1\right)
\end{equation}

\section{Qualitative Analysis and Scenario Evaluation}

Evaluating anonymization performance across complex Kuala Lumpur traffic imagery requires analyzing qualitative behavior across challenging operational scenarios encountered in the dataset.

\subsection{Qualitative Analysis across Traffic Scenarios}

\begin{figure*}[h!]
\centering
\subcaptionbox{Scenario 1: High-density motorcycle stream with helmeted riders.\label{subfig:sc1}}{
    \includegraphics[width=0.45\textwidth]{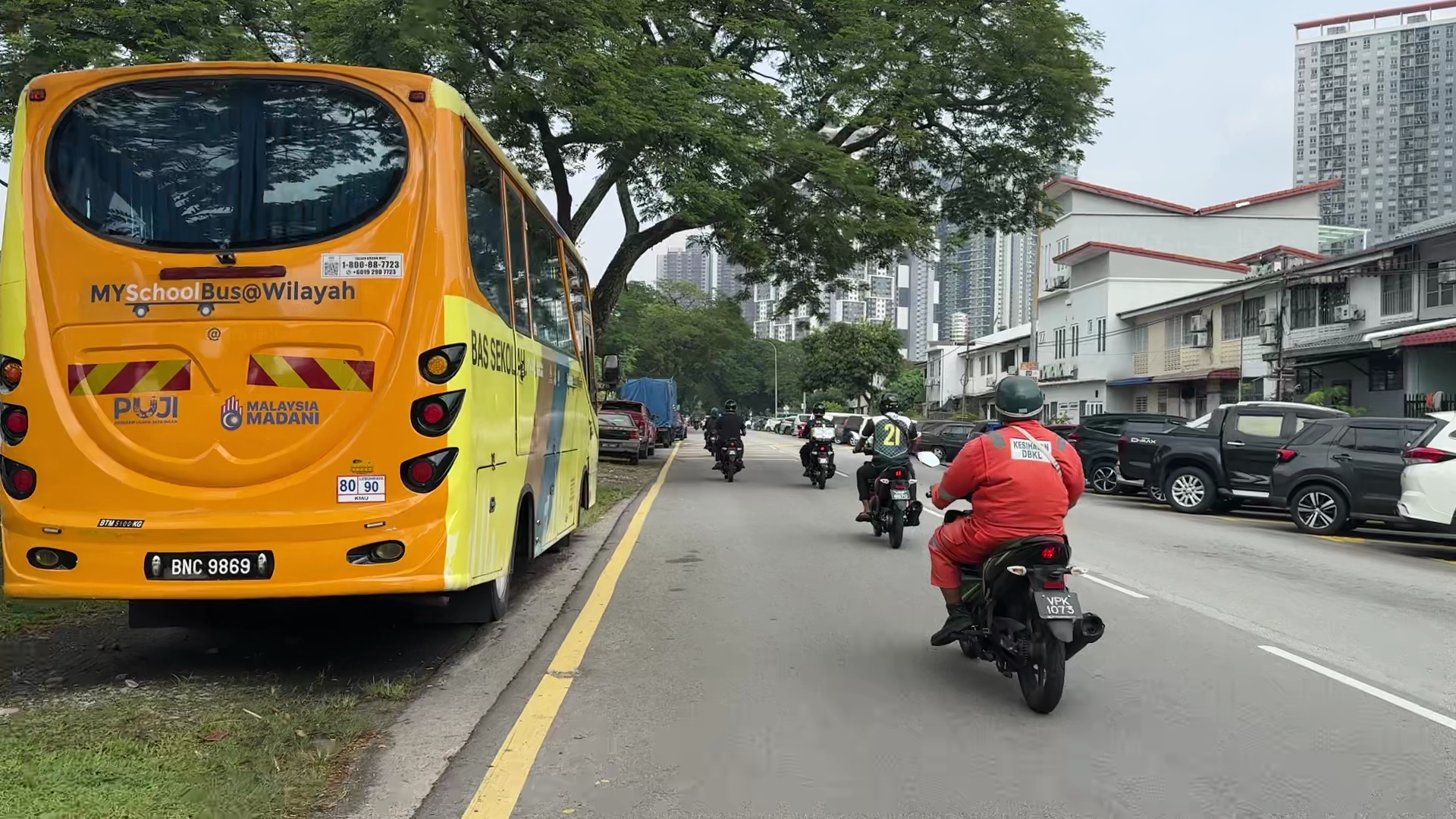}
    \hspace{0.04\textwidth}
    \includegraphics[width=0.45\textwidth]{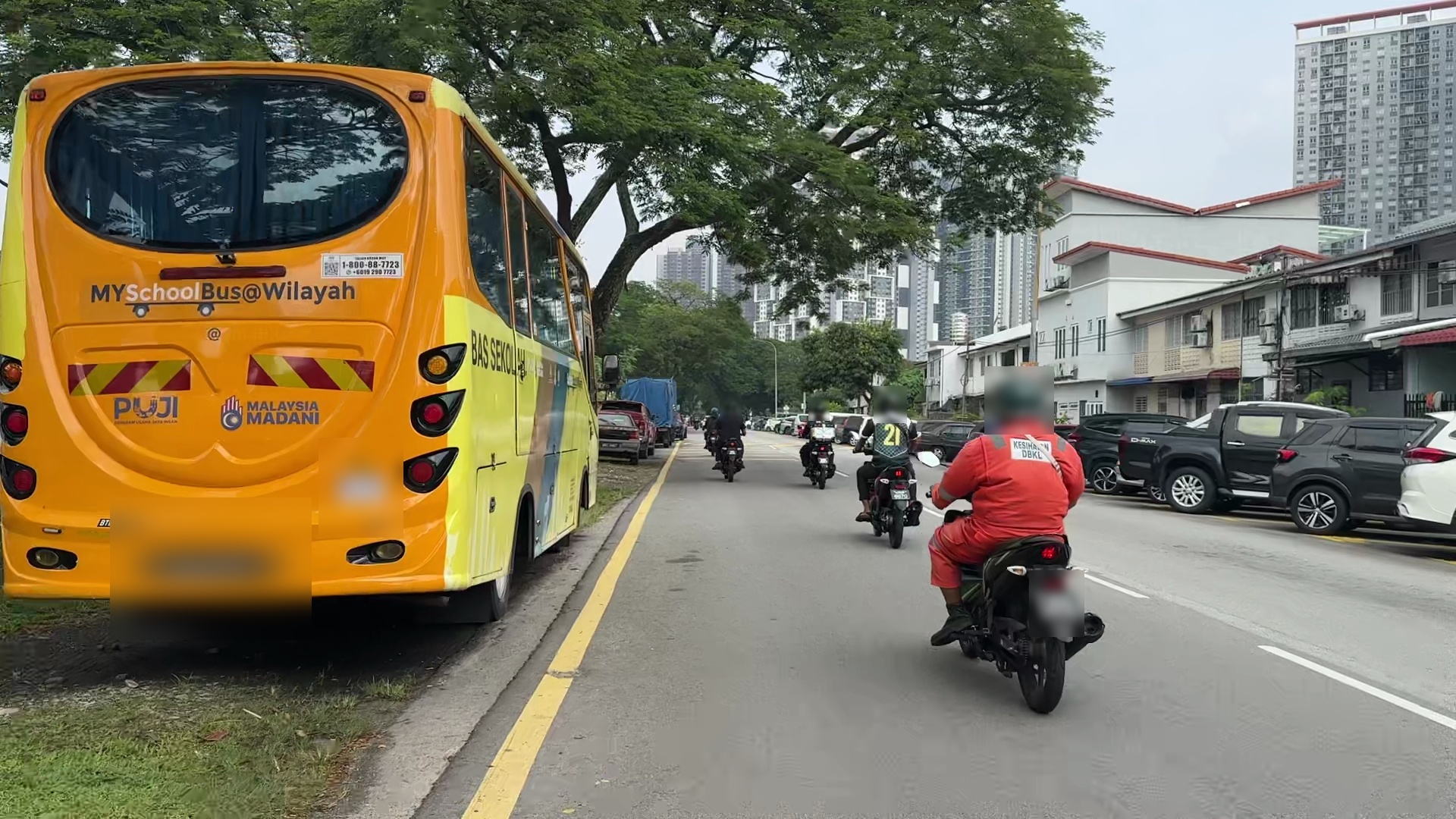}
}
\vspace{0.4cm}

\subcaptionbox{Scenario 2: Background false positive suppression via vehicle ROI containment.\label{subfig:sc2}}{
    \includegraphics[width=0.45\textwidth]{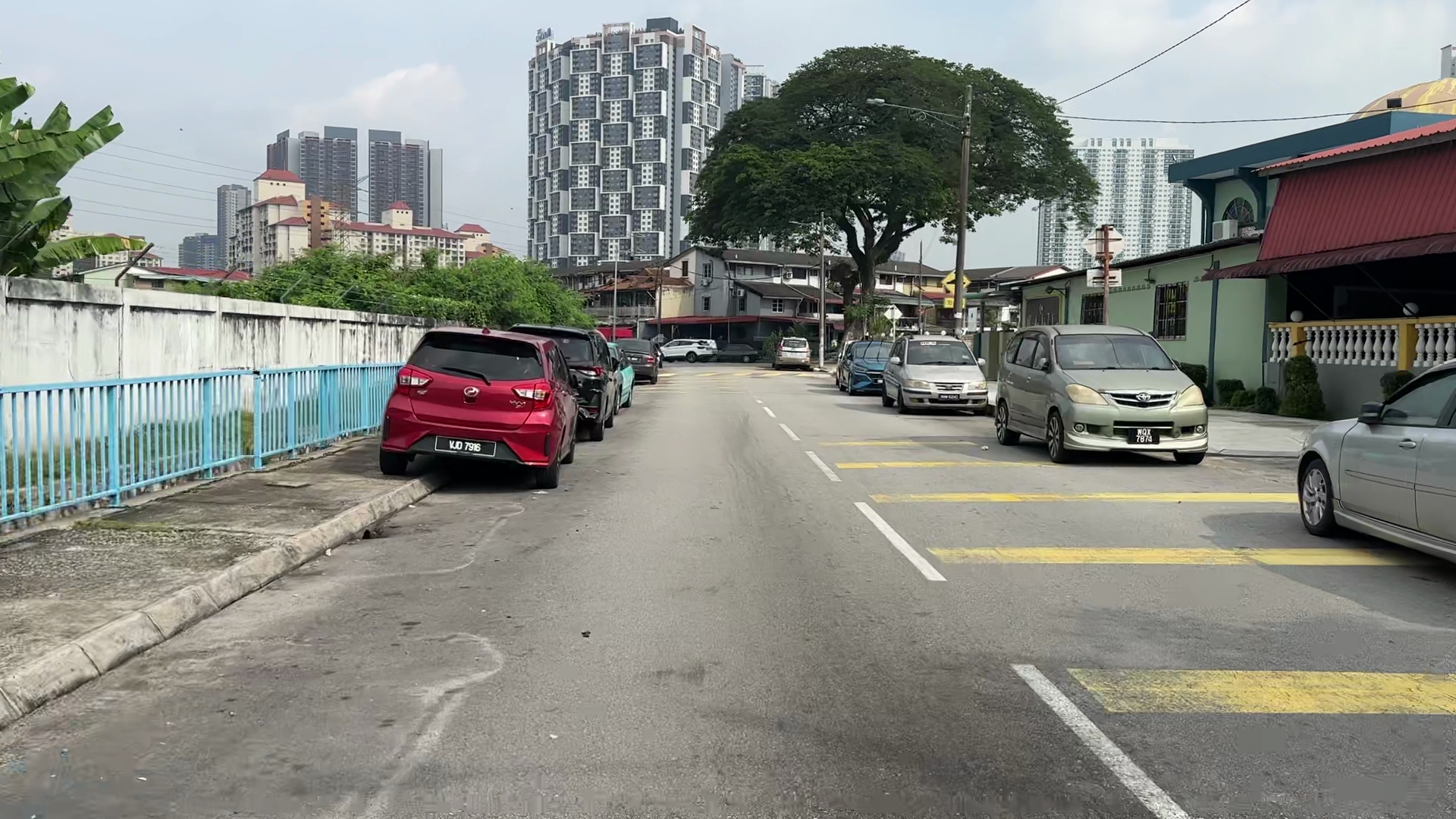}
    \hspace{0.04\textwidth}
    \includegraphics[width=0.45\textwidth]{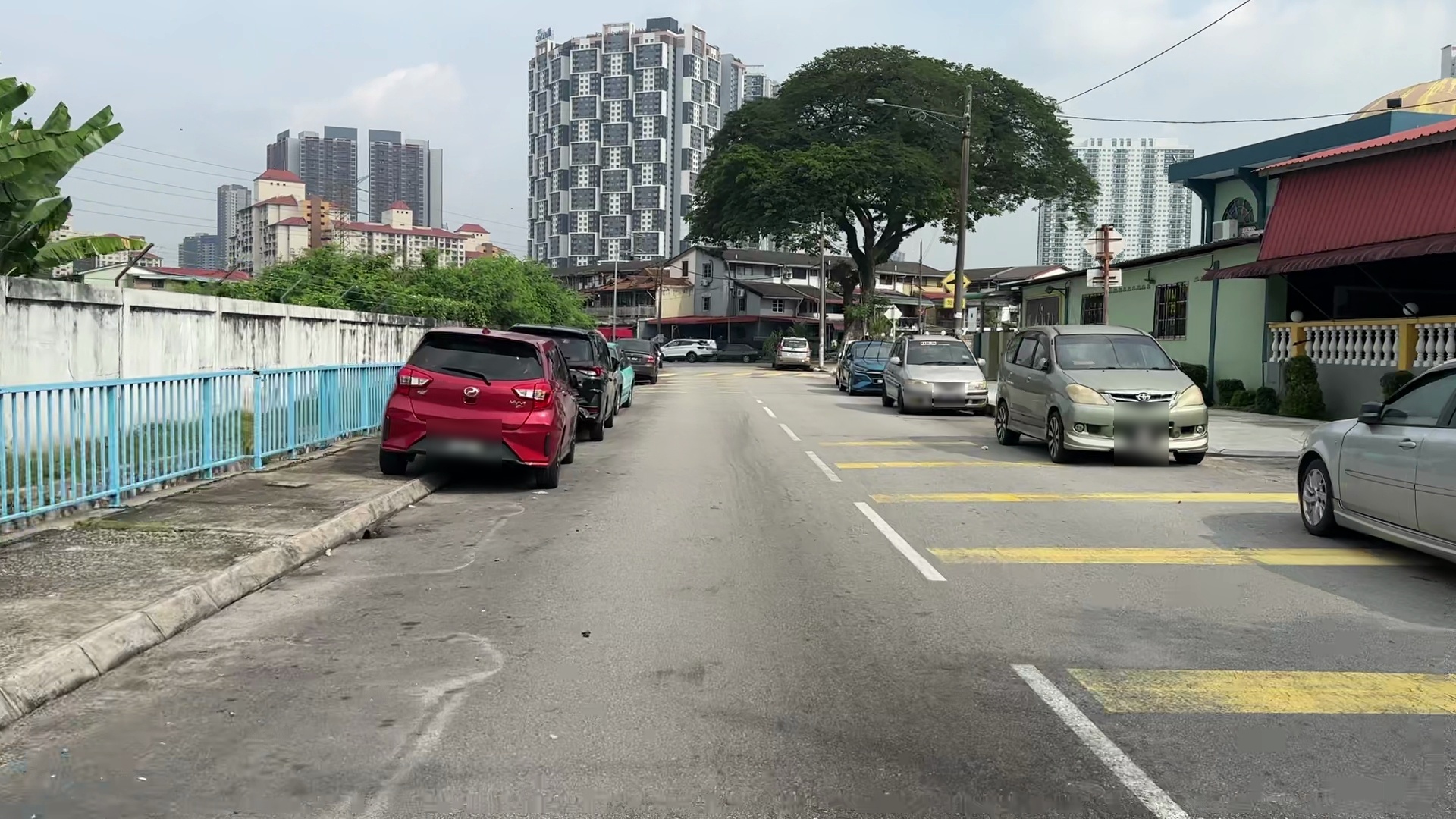}
}
\vspace{0.4cm}

\subcaptionbox{Scenario 3: Angled vehicle plates under dynamic bicycle camera roll.\label{subfig:sc3}}{
    \includegraphics[width=0.45\textwidth]{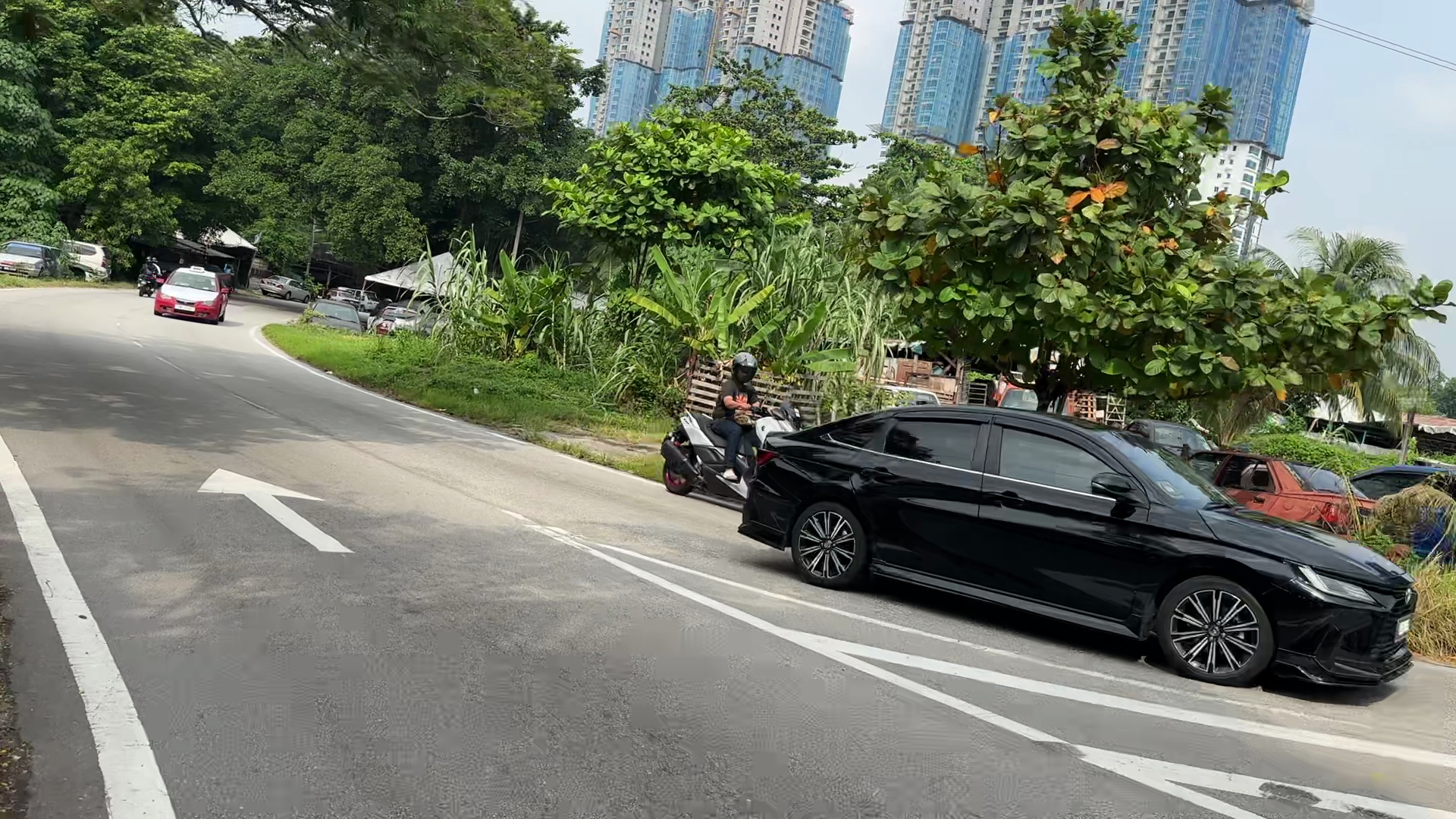}
    \hspace{0.04\textwidth}
    \includegraphics[width=0.45\textwidth]{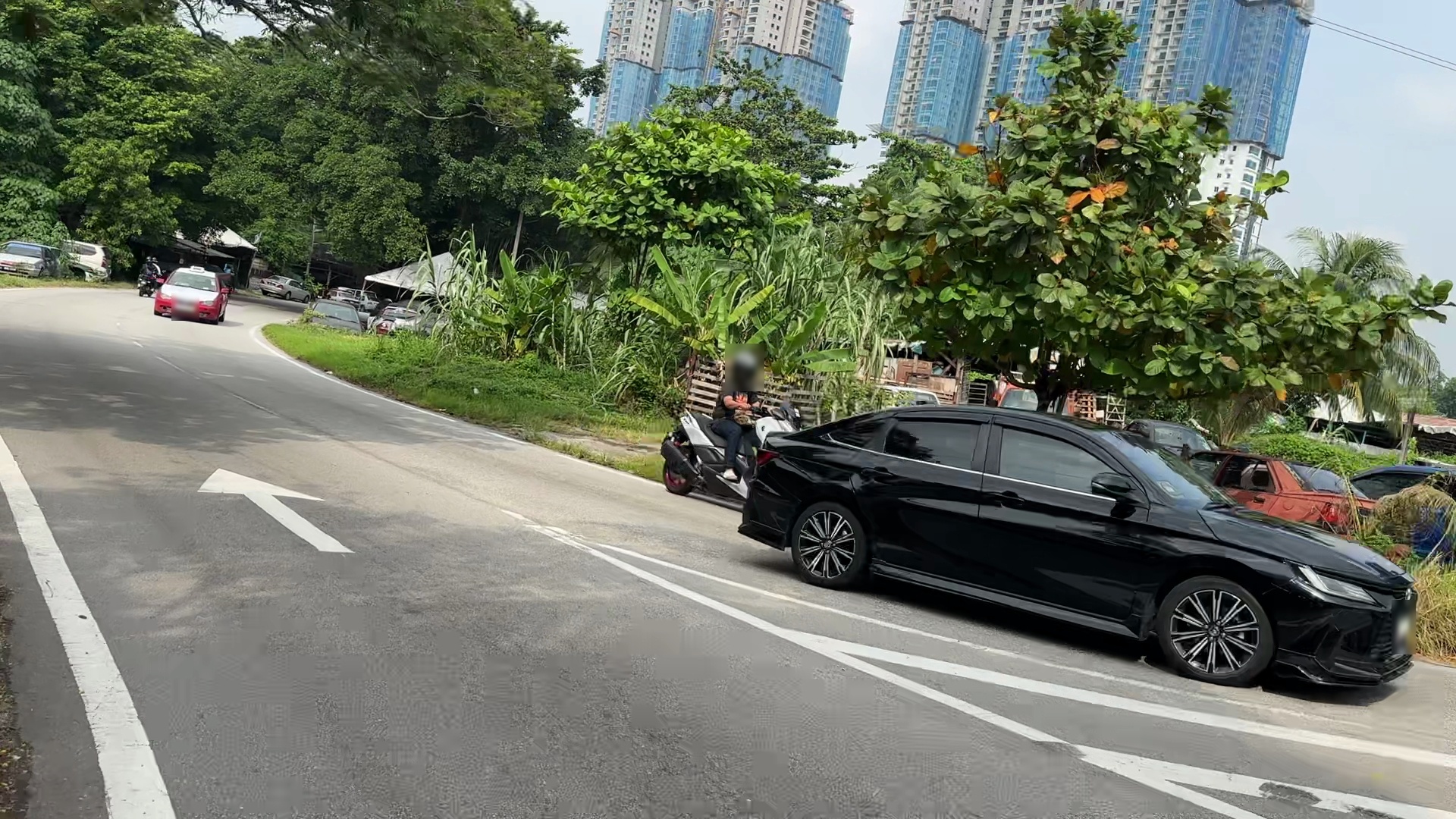}
}
\caption{Qualitative anonymization results on raw extracted frames (left) versus obfuscated outputs (right) from the Kuala Lumpur Road Dataset. Row (a): Helmeted motorcyclists and oblique tail plates. Row (b): Background noise rejection. Row (c): Rotated plates under sensor motion.}
\label{fig:qualitative_results}
\end{figure*}

Figure \ref{fig:qualitative_results} illustrates qualitative outputs generated by the proposed framework across key challenge scenarios in Kuala Lumpur traffic.

\begin{enumerate}
\item \textbf{Scenario 1: Dense Motorcycle Traffic and Helmeted Riders:} In raw imagery capturing dense lane-filtering motorcycles, conventional face detectors fail because riders wear full-face helmets or face away from the camera. Legacy plate detectors miss small, oblique tail-mounted plates. By prompting Grounding DINO \cite{liu2023grounding} with \textit{``head, helmet, motorcycle license plate''}, the pipeline captures helmeted rider profiles and angled rear plates simultaneously.
\item \textbf{Scenario 2: High-Contrast Background Noise Suppression:} Specular reflections and high-contrast street elements often trigger false positive activations in unconstrained object detectors. The Spatial Vehicle ROI Containment Engine evaluates the centroid $\mathbf{c}_p$ of each candidate plate box and verifies whether it lies within an expanded vehicle boundary $\mathbf{b}_v^\epsilon$. Detections outside vehicle hulls are discarded, preserving surrounding environmental context.
\item \textbf{Scenario 3: Dynamic Bicycle Camera Roll and Angled Plates:} Cycling motion generates variable pitch and roll ($\phi, \theta$), causing observed license plates to appear at acute rotational angles. Cross-modal attention maps in Grounding DINO attend to global plate boundaries regardless of axis alignment, while 25\% horizontal mask expansion guarantees full character coverage.
\end{enumerate}

\section{Temporal Persistence and Quality Control Auditing}

\subsection{Temporal Persistence Tracking}

To bridge single-frame detection dropouts in extracted 2 FPS video sequences, the framework executes a temporal linear interpolation pass. For frame sequence $\mathbf{I}_{t-1}, \mathbf{I}_t, \mathbf{I}_{t+1}$, if a target box is detected at $t-1$ ($\mathbf{b}_{t-1}$) and $t+1$ ($\mathbf{b}_{t+1}$) but missing at frame $t$, the missing coordinate is interpolated:

\begin{equation}
\mathbf{b}_t = \frac{\mathbf{b}_{t-1} + \mathbf{b}_{t+1}}{2}
\end{equation}

\subsection{Automated Quality Control (QC) Audit Architecture}

To eliminate manual review across thousands of frames, an automated context-aware auditor evaluates two spatial consistency rules using a large secondary context model (YOLOv8l \cite{yolov8_ultralytics}):

\begin{enumerate}
\item \textbf{Vehicle-Plate Audit Rule:} If a vehicle box $\mathbf{b}_v$ occupies an area $\ge 3000 \text{ pixels}^2$ but contains no validated plate mask ($\mathbf{M} \cap \mathbf{b}_v^\epsilon = \emptyset$), the frame is flagged.
\item \textbf{Person-Face Audit Rule:} If a pedestrian/rider box $\mathbf{b}_{\text{person}}$ exceeds a height threshold $\ge 40 \text{ pixels}$ but contains no face mask ($\mathbf{M} \cap \mathbf{b}_{\text{person}} = \emptyset$), the frame is flagged.
\end{enumerate}

Passed frames route directly to \texttt{high\_confidence/}, while flagged frames isolate into \texttt{review\_required/} alongside an automated audit summary (saved as \texttt{qc\_summary\_report.json}). This context audit reduces manual review requirements significantly, providing a defensible audit trail for privacy compliance under legal frameworks like the Malaysian PDPA 2010 and GDPR.

\section{Conclusion and Code Availability}

Curating open-source urban traffic datasets in complex tropical environments requires specialized computer vision pipelines capable of handling non-standard vehicle modifications, high motorcycle density, and dynamic sensor motion. We demonstrated that traditional rigid object detectors and standard CNNs are insufficient for the extreme variability of Kuala Lumpur traffic imagery. By pairing Grounding DINO vision-language transformers with a Spatial Vehicle ROI Containment Engine, our framework achieves accurate anonymization of faces, helmeted heads, and custom license plates while suppressing environmental false positives.

To support reproducible research and open science, the complete source code, spatial ROI engine implementation, and interactive demonstration notebook (.ipynb) are open-sourced at \url{https://github.com/TanzinAbdul/Kuala-Lumpur-Road-Dataset-Anonymizer}.

\bibliographystyle{elsarticle-num}
\bibliography{references}

\end{document}